%% file: main.tex
\documentclass[letterpaper, 10 pt, conference]{IEEEtran}
\IEEEoverridecommandlockouts

\input{header}

\begin{document}

    \title{
        StaR Maps: Unveiling Uncertainty in Geospatial Relations
    }

    \author{
        Benedict Flade$^{1, *}$, Simon Kohaut$^{2, *}$, Julian Eggert$^{1}$, Devendra Singh Dhami$^{3}$, Kristian Kersting$^{2, 4, 5, 6}$
        \thanks{
            $^{*}$ Authors contributed equally
        }%
        \thanks{
            $^{1}$ Honda Research Institute Europe GmbH, \newline\hspace*{1.6em} 
            Carl-Legien-Str. 30, 63073 Offenbach, Germany \newline\hspace*{1.6em}
            {\tt\small fist.last@honda-ri.de}
        }%
        \thanks{
            $^{2}$ Artificial Intelligence and Machine Learning Lab, \newline\hspace*{1.6em} 
            Department of Computer Science, \newline\hspace*{1.6em}
            TU Darmstadt, 64283 Darmstadt, Germany
        }%
        \thanks{
            $^{3}$
            Uncertainty in Artificial Intelligence Group, \newline\hspace*{1.6em}
            Department of Mathematics and Computer Science, \newline\hspace*{1.6em}
            TU Eindhoven, 5600 MB Eindhoven, Netherlands%
        }%
        \thanks{
            $^{4}$ Hessian AI
        }%
        \thanks{
            $^{5}$ Centre for Cognitive Science
        }%
        \thanks{
            $^{6}$ German Center for Artificial Intelligence (DFKI)
        }%
    }

    \maketitle

    \begin{abstract}
        \input{content/0_abstract}
    \end{abstract}
    \IEEEpeerreviewmaketitle

    \input{content/1_introduction}
    \input{content/2_related_work}
    \input{content/3_methods}
    \input{content/4_experiments}

    \input{content/5_conclusion}
    \input{content/6_acknowledgment}

    \bibliographystyle{IEEEtran}
    \bibliography{references.bib}
\end{document}

%% file: header.tex
\usepackage{cite}
\usepackage{amsmath,amssymb,amsfonts}
\usepackage{algorithmic}
\usepackage{graphicx}
\usepackage{textcomp}
\usepackage[dvipsnames]{xcolor}
\usepackage{balance}
\usepackage{hyperref}
\usepackage{epstopdf}
\usepackage[normalem]{ulem}
\usepackage{upgreek}
\usepackage{comment}
\usepackage{url}
\usepackage{tikz}
\usepackage{lipsum}
\usepackage[formats]{listings}
\usepackage{subcaption}
\usepackage{multirow}
\usepackage[frozencache,cachedir=.]{minted}


%% file: content/0_abstract.tex
The growing complexity of intelligent transportation systems and their applications in public spaces has increased the demand for expressive and versatile knowledge representation.
While various mapping efforts have achieved widespread coverage, including detailed annotation of features with semantic labels, it is essential to understand their inherent uncertainties, which are commonly underrepresented by the respective geographic information systems.
Hence, it is critical to develop a representation that combines a statistical, probabilistic perspective with the relational nature of geospatial data.
Further, such a representation should facilitate an honest view of the data's accuracy and provide an environment for high-level reasoning to obtain novel insights from task-dependent queries.
Our work addresses this gap in two ways.
First, we present Statistical Relational Maps (StaR Maps) as a representation of uncertain, semantic map data.
Second, we demonstrate efficient computation of StaR Maps to scale the approach to wide urban spaces.
Through experiments on real-world, crowd-sourced data, we underpin the application and utility of StaR Maps in terms of representing uncertain knowledge and reasoning for complex geospatial information.


%% file: content/1_introduction.tex
\section{Introduction}
\label{sec:introduction}

\input{figures/motivation}

As intelligent transportation systems become increasingly sophisticated, we must carefully represent the agent's environment and accurately localize relevant features.
Such systems fundamentally rely on environment representations that offer geometrical and relational information, ranging from static entities like roads to dynamic elements such as other traffic participants and pedestrians.
Modern maps have been shown to include detailed 3D point clouds and semantic maps marked by polylines, such as road lane markings. 
Such maps serve diverse purposes and have been refined to meet nearly every accuracy requirement, provided that the hardware and conditions are adequate.
For example, advancements in localization technologies, such as Real-Time Kinematics with Global Navigation Satellite Systems, have significantly improved accuracy in map creation.
However, obtaining high-quality map data over large areas, such as entire cities or regions, can quickly become an intractable task due to the cost of the equipment and frequent changes in the respective areas.
Instead, today's and future systems must work with data sources of mixed levels of accuracy, i.e., from crowd-sourced to high-definition, while incorporating their noisy perception of the environment.

Traditional paper maps often include valuable accuracy statements such as diagrams and estimates of horizontal and vertical positional errors. 
Regrettably, such uncertainty information is frequently absent in the digital age~\cite{GHunter1999}. 
When drawing conclusions from map data of multiple sources, we believe that an understanding of the accuracy of each individual map element can significantly enhance spatial reasoning by providing uncertainty information (see Figure~\ref{fig:motivation}). 

However, with digital maps offering entirely new possibilities for data management, more than merely including an accuracy diagram is required.
Besides categorical and quantitative uncertainties, it is important to facilitate reasoning over the data that attends to the attached uncertainties. 
For example, consider a scenario where the permitted driving speed is related to the distance to the next junction.
With uncertainty of the agent's location and the exact location of the junction, it is important to provide a probabilistic perspective into the agent's decision-making process.

Nevertheless, the integration of detailed statistical data into maps has been limited. 
This oversight may be due to the previously adequate levels of accuracy for the needs of most systems or the focus on meeting specified accuracy requirements rather than the need to analyze and evaluate its uncertainty. 
However, with the combination of statistical and probabilistic approaches being on the rise~\cite{Deraedt2016}, it is appropriate to consider whether and how environmental representation should be adapted to benefit from this.

To this end, we introduce Statistical Relational Maps (StaR Maps). 
StaR Maps provide hybrid probabilistic and spatial relations for navigation and demonstrate their broad range of applications in designing a novel probabilistic framework for planning and complex geospatial queries. 

Our contributions are two-fold:
\begin{itemize}
    \item We present Statistical Relational Maps (StaR Maps), hybrid probabilistic, semantic maps that embrace mixed levels of uncertainty in geographic features and provide probabilistic answers to spatial queries.
    \item We demonstrate how StaR Maps underpin methods that rely on interpretable semantics to constrain and describe navigation tasks, i.e., utilizing a rich vocabulary of spatial relations that StaR Maps provide.
\end{itemize}
Our results are published as Open Source implementation within our Probabilistic Mission (ProMis) framework at \url{https://github.com/HRI-EU/ProMis}.

%% file: figures/motivation.tex
\begin{figure}
    \centering
    \includegraphics[width=0.9\linewidth]{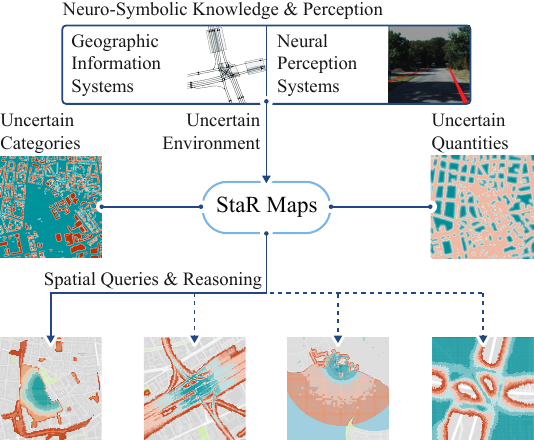}
    \caption{
        \textbf{Statistical Relational Maps (StaR Maps) capture uncertain environments:}
        StaR Maps provide a unified interface to heterogeneous background knowledge.
        Given symbolic, geographic data, and neural perception of the navigation space, they offer semantic, probabilistic answers to first-order logic based spatial queries.
    }
    \label{fig:motivation}
\end{figure}

%% file: content/2_related_work.tex
\input{figures/architecture}

\section{Related Work}
\label{sec:related_work}
\subsection{Probabilistic Localization} 
\label{subsec:prob_loc}

In this work, we address the topic of uncertainty information in the context of geospatial data. 
One crucial aspect of this is the topic of localization, i.e., estimating the own or another traffic participant's position and attitude.
In absolute localization systems, i.e., with regard to a global reference, approaches have progressed from conventional Global Navigation Satellite Systems (GNSS) to more sophisticated methods such as Real-Time Kinematic (RTK) GNSS~\cite{Ando2021}.
Methods that address challenges in absolute localization, e.g., multipath effects in urban environments, can improve localization accuracy significantly~\cite{Dong2016}.
Similarly, proprioceptive sensors observing internal states, e.g., acceleration~\cite{Nilsson2014}, or exteroceptive sensors observing the environment, e.g., cameras~\cite{Alkendi2021}, LiDAR~\cite{Demir2019}, or radar~\cite{Ward2016}, are used to improve localization accuracy.
It has been shown that the ego-position, e.g., for street vehicles, can be improved by comparing proprioceptive and exteroceptive observations with features stored in a map~\cite{OPink2008, ASchindler2013}.

For map-relative localization, environment representations range from 3D point clouds~\cite{Caselitz2016} to polyline-level road shapes~\cite{Flade2017}, with road markings presenting one of the most commonly used entities of reference~\cite{Lu2017, Flade2018}. 
Furthermore, reaching resource requirements can be achieved by choosing a low-cost approach targeting affordable sensor setups~\cite{Flade2020}.
However, understanding uncertainties in the underlying data is equally crucial for quantifying the localization system's output's trustworthiness. 
For example, localization approaches based on probabilistic methods, e.g., Bayes filters such as variations of the Kalman filter~\cite{Rezaei2007, Ballardini2017}, provide confidence data alongside the expected position information. 

\subsection{Statistical Relational Artificial Intelligence}

Researchers and engineers have considered various approaches to meet the demands of intelligent transportation systems.
To this end, both reasoning, e.g., based on propositional or first-order logic, have been successfully employed for representing and understanding the environment and task at hand.
The emerging field of Statistical Relational Artificial Intelligence (StaR AI) has shown meaningful advances in combining tools of those fields into more robust and expressive inference systems \cite{marra2024statistical}.

StaR AI approaches have in the past been built based on programmatic reasoning systems in first-order logic such as Prolog~\cite{colmerauer1990introduction}.
That is, probabilistic extensions that embrace uncertainties in formal logic, e.g., Bayesian Logic Programs~\cite{bayesian_logic} and Probabilistic Logic Programs~\cite{problog,inference_in_plp}, allow for probabilistic inference in first-order logic.
While they were not formulated for end-to-end learning with artificial neural networks, Neuro-Symbolic models such as DeepProbLog~\cite{deepproblog}, NeurASP~\cite{neurasp} and SLASH~\cite{slash} have been introduced to close this gap, embracing the advantages of neural information processing and probabilistic reasoning into a single framework.

Applying these methods in building autonomous agents, e.g., robots acting in logically constrained environments, is attractive to ensure safe and reliable decision-making.
Notably, systems such as ProMis~\cite{Kohaut2023} utilize hybrid probabilistic logic~\cite{nitti, kumar2023first}, combining categorical and continuous distributions into a spatially independent and identically distributed navigation knowledge and constraints model.
As a result of this, such systems provide an interpretable and adaptable framework for planning tasks.

\subsection{Uncertainty-Aware Environment Representation}

Two major requirements must be met for intelligent transportation systems to leverage the strengths of current and future StaR AI approaches in spatial tasks, such as laying out optimal paths or searching for spaces that fulfill predetermined requirements.
First, the logical component of statistical relational approaches is based on symbols and semantics that allow for modeling entities, their relations, and interactions.
Transferred to a representation of the environment, features of the environment need to be assigned precise semantics.

While, e.g., unstructured point cloud maps or unlabeled structures in image data are unsuitable, segmentation approaches such as segment anything~\cite{Kirillov2023} and object-centric vision~\cite{locatello2020object,kipfconditional} can help to separate objects which can be assigned semantic labels.
To this end, work on multi-target tracking~\cite{chen2019multi} and point cloud classifiers~\cite{Himmelsbach2010} have shown successful identification and tracking of objects with their assigned semantics.  

A prominent example of creating a semantically annotated, wide-coverage environment representation is the crowd-sourced project OpenStreetMap~\cite{Haklay2008}, storing its geospatial data in nodes and ways alongside a rich tagging system.
Further examples of semantic maps include Lanelets~\cite{Bender2014}, atomic interconnected road segments that may carry additional data to describe the static environment.
Additionally, approaches such as the Relational Local Dynamic Map~\cite{Eggert2017} represent the environment's static, quasi-static, transient, and dynamic entities as a holistic, interconnected graph.

Many map-creation methods exist. 
From Lidar- and camera-based mapping~\cite{Levinson2010, Huang2019a, Lategahn2012} via map data derived from aerial imagery~\cite{Wei2022}, to map generation by analyzing GNSS-based vehicle trajectories~\cite{BetailleToledo2010a}.
When employing such localization approaches to create maps, the uncertainties of the map-making process also apply to the generated map.
In the worst case, the quality of geospatial data in an area, e.g., within a set of local roads, may drastically differ from feature to feature.
Even within a single framework such as OpenStreetMap, data from amateur cartographers with affordable equipment intersects with professional data from commercial navigation information providers.
More specifically, the authors of~\cite{Haklay2008} criticize that OpenStreetMap does not provide information on the data's positional accuracy.
Furthermore, being based on multiple heterogeneous data sources that have extensive coverage introduces not only different variances but distinct error types, and how to account for them simultaneously in a formalized holistic and map-centered way needs to be addressed.

To bridge this gap, we introduce StaR Maps, a novel environment representation and geographic reasoning system that considers heterogeneous sources of map data by incorporating different levels of fine-granular spatial uncertainty into a unified, semantically annotated environment representation.
The novelty is a holistic framework that allows the representation of map-related error parameters, i.e., uncertainty information, processes this information, and stores this processed data in view of being subsequently used by applications that rely on a clear understanding of data quality. 
Hence, StaR Maps allows for a probabilistic worldview and facilitates complex, spatial reasoning and inference about the application and domain-specific queries. 

%% file: figures/architecture.tex
\begin{figure*}
    \centering
    \includegraphics[width=\textwidth]{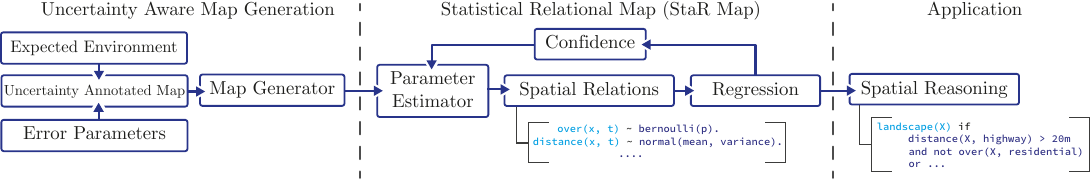}
    \caption{
        \textbf{The Statistical Relational Maps (StaR Maps) architecture:}
        Uncertain maps, annotated with translational and transformational statistics, are passed into a density estimator.
        Given a set of sample locations and the map sampler, generating variations of the annotated data according to its uncertainty, spatial relations are estimated.
        Using, e.g., moment matching, for each sample request the spatial relations are represented as categorical or continuous distributions.
        Finally, a scalar field for each relation is interpolated or approximated from the samples, providing a basis for spatial reasoning.
    }
    \label{fig:architecture}
\end{figure*}

%% file: content/3_methods.tex
\section{Methods}
\label{sec:methods}

This section introduces the concept of Statistical Relational Maps (StaR Maps). 
As motivated before, StaR Maps provide an approach to represent and reason on uncertain spatial, relational information. 
We first define how to combine expectations about the environment with a stochastic error model into Uncertainty Annotated Maps and how they facilitate sampling spatial relations, functions that relate points in navigation space with semantic map features. 
By further dividing spatial relations into categorical and quantitative relations, e.g., the predicate of a point being over a type of feature or its distance from it.
Finally, we show how StaR Maps choose support points to create scalar fields for each of the spatial relations' parameters.

\subsection{Uncertainty Aware Map Generation}
\label{sec:uam}

Uncertainty information is crucial when interpreting map data or using it to reason about spatial tasks such as navigation. 
However, before discussing the implications and utility of considering uncertainty in maps, let us define what we will understand as a map in the following.\\

\textbf{Definition 1.1 (Map).}
A Map $\mathcal{M} = (\mathcal{V}, \mathcal{E}, \tau)$ is a triple of vertices $\mathcal{V}$, edges $\mathcal{E}$ and tagging function $\tau$.
Each vertex $\vec{v}$ is in a $d$-space $\mathbb{R}^d$ of Cartesian coordinates.
If a path exists between two vertices in $\mathcal{V}$ across edges in $\mathcal{E}$, they are part of the same \textit{feature}.
For each vertex $\vec{v} \in \mathcal{V}$, the function $\tau(\vec{v}) \subseteq \mathbb{T}$ annotates $\vec{v}$ with a set of tags, providing semantics such as road or building types.\\

Today, many examples of available maps exist, e.g., the crowd-sourced OpenStreetMap~\cite{Haklay2008} or concepts such as the Relational Local Dynamic Map\cite{Eggert2017}.
While these concepts can represent an expectation, they commonly lack the associated uncertainty data.
As depicted in Figure~\ref{fig:architecture}, we are now introducing uncertainty-aware map generation, combining expectations of the environment represented as a Map with an error model, describing how individual vertices or features likely deviate from the expectation due to the employed sensing methods and equipment.
This combination of map data and uncertainty parameters leads us to the following definition.\\

\textbf{Definition 1.2 (Uncertainty Annotated Map).}
An Uncertainty Annotated Map (UAM) $\mathcal{U} = (\mathcal{M}, a, b)$ is a triple of a map $\mathcal{M}$ and two annotator functions $a$ and $b$.
While the map follows Definition~1.1., functions $a(\vec{v}) = \vec{\alpha}$ and $b(\vec{v}) = \vec{\beta}$ assign each vertex $\vec{v}$ transformation parameters $\vec{\alpha}$ and translation parameters $\vec{\beta}$. \\

Here, $\vec{\alpha}$ and $\vec{\beta}$ are employed as statistical moments.
For example, $\vec{\beta}$ might contain mean and variance of a randomly distributed translation.
Hence, a UAM represents expectations alongside the parameters of a stochastic error model.
While annotating each vertex of a map with individual parameters can be beneficial rather than generating entire map features using the same spatial uncertainty, this choice in granularity will have no systematic impact on our further discussions.

A UAM can be employed as generator of likely maps, e.g., facilitating how we will sample data from spatial relations in Section~\ref{sec:spatial_relations}. 
To this end, we apply the following affine map as a simple error model, covering perturbations such as translations, rotation, scaling, or shearing effects.
Figure~\ref{fig:map_variations} illustrates this process and how, e.g., a distance from a fixed point to the closest road changes across sampled maps.

Similar to prior work~\cite{flade2021error}, we consider for each $\vec{v} \in \mathcal{V}$ of map $\mathcal{M}$ the following manipulations: 
\begin{align*}
    \vec{\Phi}_n &\leftarrow \phi(\vec{\alpha}) = \phi(a(\vec{v})) & \quad \quad \text{(Transformation)} \\
    \vec{t}_n &\leftarrow \kappa(\vec{\beta}_i) = \kappa(b(\vec{v})) & \quad \quad \text{(Translation)} 
\end{align*}
Here, $\phi$ generates matrices $\vec{\Phi}_n$ to apply geometric transformations, e.g., rotations, that keep the center point fixed while $\kappa$ generates offset vectors $\vec{t}_n$ to apply a translation.
Hence, from an original vertex $\vec{v}$ we can generate
\begin{align}
\vec{v}_n &= \vec{\Phi}_n \cdot \vec{v} + \vec{t}_n
\end{align}
We are repeating this process across the map $N$ times, resulting in a collection $\mathcal{W} = \{\mathcal{M}_0, ..., \mathcal{M}_N\}$.
In the following, we employ $\mathcal{W}$ to estimate the parameters of StaR Maps' spatial relations.

\input{figures/map_variations}
\input{figures/distributional_atoms}

Note that the proposed error models allow the user to make assumptions about the error, e.g., considering how the map is generated.
Such uncertainty information could be derived from the sensors used to create the maps, as far as such information is available. 
However, in cases with limited prior knowledge, we previously showed how to identify such error parameters online for road segments by observing deviations with a hybrid localization system~\cite{flade2021error}.

\subsection{Spatial Relations}
\label{sec:spatial_relations}

Spatial relations are crucial for constraining the operation of intelligent transportation systems, whether on the ground, in the air, or submerged in water.
Regardless of application, adhering to precise rules in an uncertain environment is a common requirement.
To facilitate a framework for reasoning on such requirements, we show spatial relations as a language to express stochastic relations between points in space, types of map features, and instantiations of UAMs.

We consider the following two models of spatial relations.
\begin{align}
    f&:\ \mathbb{R}^d \times \mathbb{T} \times \mathbb{M} \rightarrow \mathbb{B} \\
    g&:\ \mathbb{R}^d \times \mathbb{T} \times \mathbb{M} \rightarrow \mathbb{R}
\end{align}
Here, a spatial relation with signature $f$ describes a categorical relationship between a point $x \in \mathbb{R}^d$, a type of map features $\tau \in \mathbb{T}$ and a map $\mathcal{M} \in \mathbb{M}$.
Analogously, spatial relations with signature $g$ describe quantitative relationships between the same input data.
Many functions will fit either definition; here, we consider the geometric relations \textit{over} and \textit{distance} as running examples of $f$- and $g$-type spatial relations, respectively.
While the former expresses whether a point in space is above or, in the 2D case, within a map feature of the referenced type, the latter encodes the distance between a point and the closest matching geometry.

Note that these are binary relations where all geospatial features of the same type are considered anonymously, i.e., as a whole. 
By doing so, rather than computing the distance between a point and all individual features, the following discussions remain computationally tractable.
Next, we discuss how the computation of spatial relations $f$ and $g$ over a set of sampled maps $\mathcal{W}$ sets up the data we need to define and apply StaR Maps.
While the differentiation between $f$ and $g$ remains important throughout the rest of this work, we will denote function $r$ as a generalization of any spatial relation.

\subsection{Statistical Relational Maps}

To summarize, we consider environment expectations alongside a stochastic error model to introduce the concept of UAMs and the process of generating a set of possible instances of maps $\mathcal{W}$ from them.
Although we have defined spatial relation $r$ to provide concrete, Boolean, or real-valued quantities, for StaR Maps, we are interested in probabilistic models that express the UAM's encoded uncertainty.
To compute these statistics for a specific spatial relation $r$ and a type $\tau$ on a point $\vec{x} \in \mathbb{R}^d$, we have to compute  $\rho_n = r(\vec{x}, \tau, \mathcal{M}_n)$ for each $\mathcal{M}_n \in \mathcal{W}$.
With the set $\mathcal{P} = \{\rho_0, ... \rho_N\}$ at hand, we can then compute the statistical moments, e.g., mean and variance of $\mathcal{P}$.
Using moment matching with the desired distribution, for example, a Bernoulli distribution for $f$- or Gaussian distribution for $g$-signature relations, we can store the resulting parameters within a StaR Map.\\

\textbf{Definition 1.3 (Statistical Relational Map).}
A Statistical Relational Map (StaR Map) $\mathcal{S} = (\mathcal{F}, \mathcal{G}, u_{i, j}, v_{k, l})$ is made up of sample collections $\mathcal{F} = \{F_{i, j}, ... F_{I, J}\}$ and $\mathcal{G} = \{G_{k, l}, ... G_{K, L}\}$ as well as the functions $u_{i, j}: \mathbb{R}^d \times \mathbb{T} \rightarrow [0, 1]^e$ and $v_{k, l}: \mathbb{R}^d \times \mathbb{T} \rightarrow \mathbb{R}^e$.
While $F_{i, j}$ and $G_{k, l}$ contain data computed for each pair of spatial relation and type over $\mathcal{W}$, the functions $u_{i, j}$ and $v_{k, l}$ approximate scalar fields for the $e$ parameters of each relation given the data in $F_{i, j}$ and $G_{k, l}$ respectively. \\

Figure~\ref{fig:distributional_atoms} illustrates the StaR Maps generation with a road network example for two spatial queries (\textit{distance} and \textit{over}) that are queried from crowd-sourced data.
A set of random maps is generated based on the UAM while considering a translational error model.
For each location in the agent's navigation frame, the parameters $\mu$ (a) and $\sigma$ (b) of a normal distribution, modeling the distance to the closest road, are computed. 
We obtain the probability of a chosen regulatory constraint being met from the cumulative distribution function of $\mathcal{N}(\mu, \sigma^2)$, or more generally via sampling. 
In this case, we chose to keep a distance of over 30 meters (c).
Finally, (d) shows the probability of a location in the agent's navigation space being occupied by buildings.

%% file: figures/map_variations.tex
\begin{figure}
    \begin{subfigure}[t]{0.15\textwidth}
        \includegraphics[width=\textwidth]{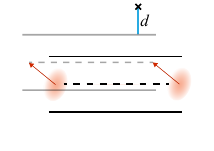}
        \caption{Translation}
    \end{subfigure}\hfill%
    \begin{subfigure}[t]{0.15\textwidth}
        \includegraphics[width=\textwidth]{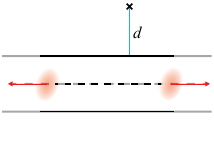}
        \caption{Scaling}
    \end{subfigure}\hfill%
    \begin{subfigure}[t]{0.15\textwidth}
        \includegraphics[width=\textwidth]{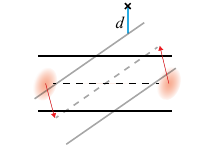}
        \caption{Rotation}
    \end{subfigure}
    \caption{
        \textbf{Sampling from UAMs:}
        Each map element carries an expectation of its true spatial occupancy, annotated with uncertainties that result from, e.g., the employed sensors or measurement methodology.
        Once a set of maps has been sampled, we probe the environment from a selection of points to fit StaR Maps' spatial relations.
        Here, sampling the distance of a point to the closest uncertain road with different error models (a - c) is illustrated.
    }
    \label{fig:map_variations}
\end{figure}

%% file: figures/distributional_atoms.tex
\begin{figure*}
    \centering
    \begin{subfigure}{0.245\textwidth}
        \includegraphics[width=\textwidth]{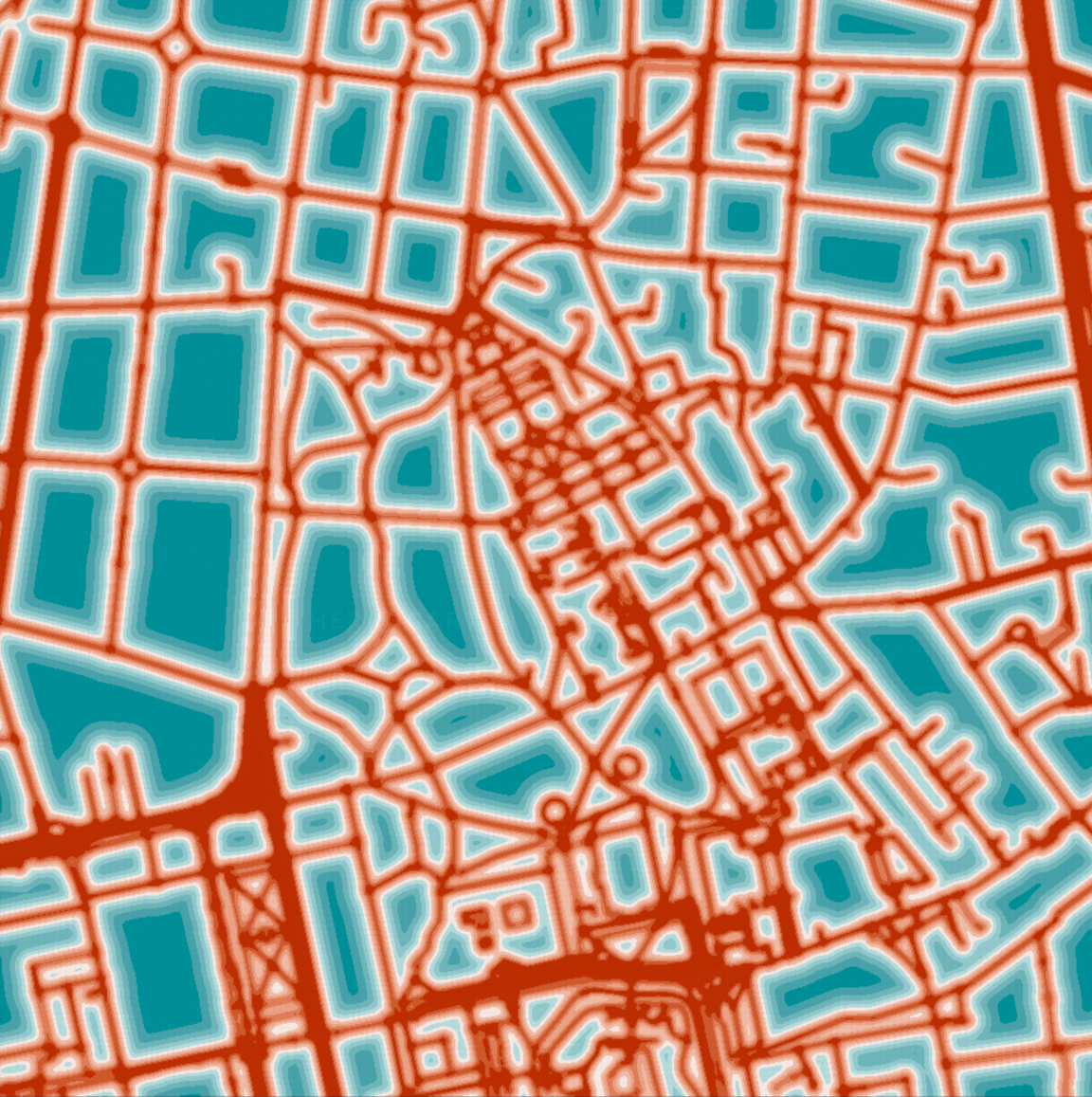}
        \caption{$\mu_D$}
    \end{subfigure}
    \begin{subfigure}{0.245\textwidth}
        \includegraphics[width=\textwidth]{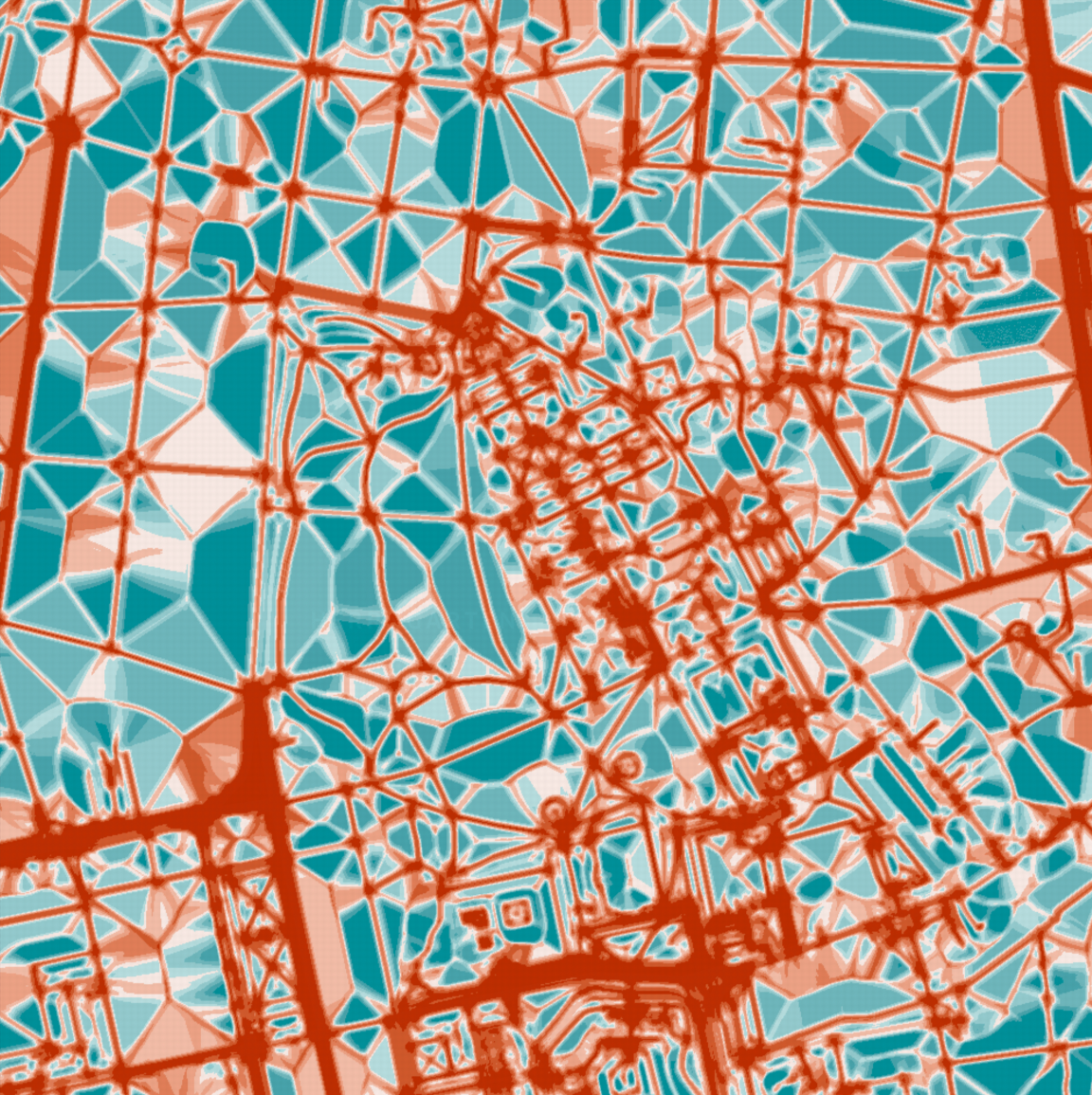}
        \caption{$\sigma^2_D$}
    \end{subfigure}
    \begin{subfigure}{0.245\textwidth}
        \includegraphics[width=\textwidth]{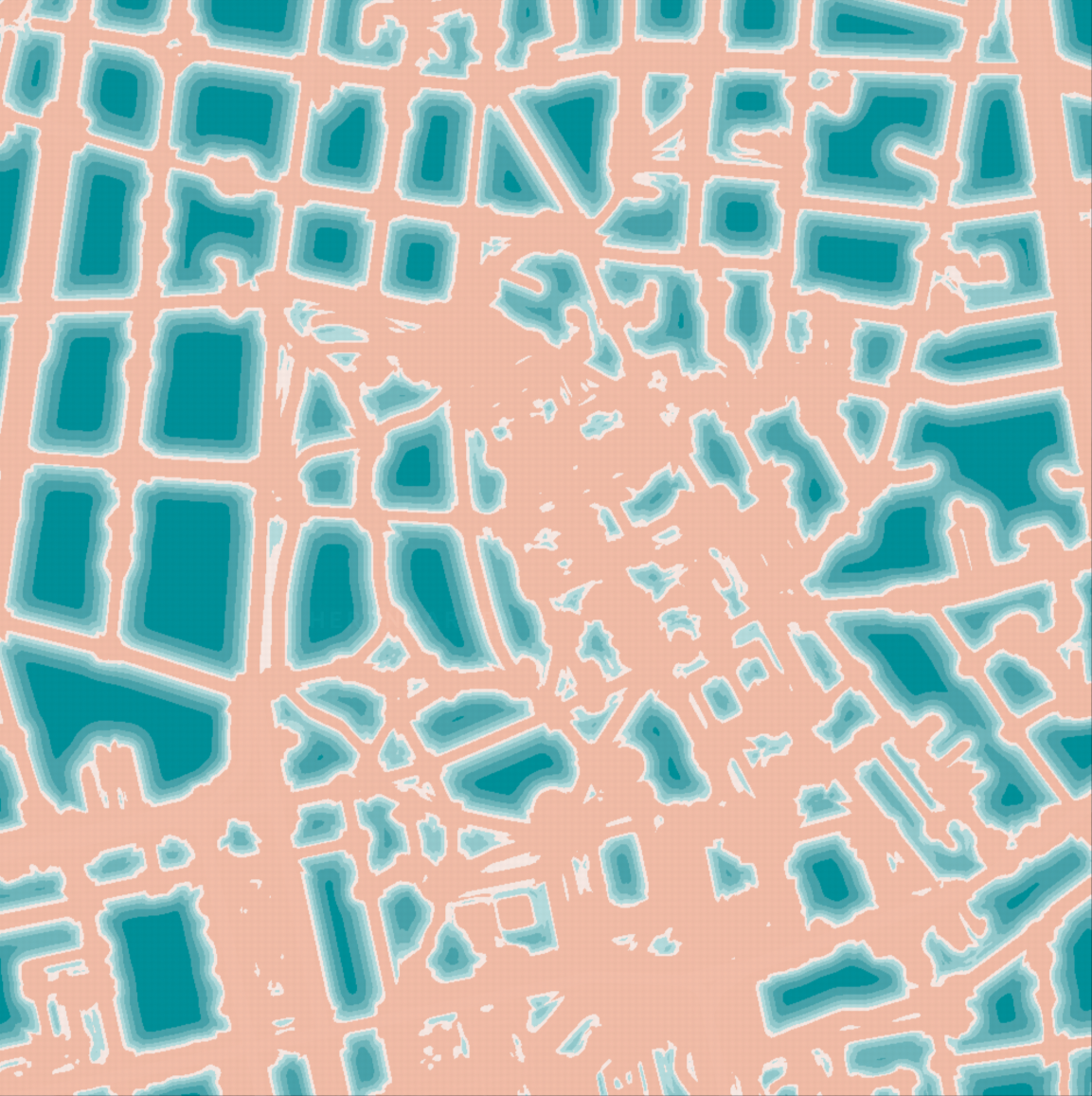}
        \caption{$P(D > 30m)$}
    \end{subfigure}
    \begin{subfigure}{0.245\textwidth}
        \includegraphics[width=\textwidth]{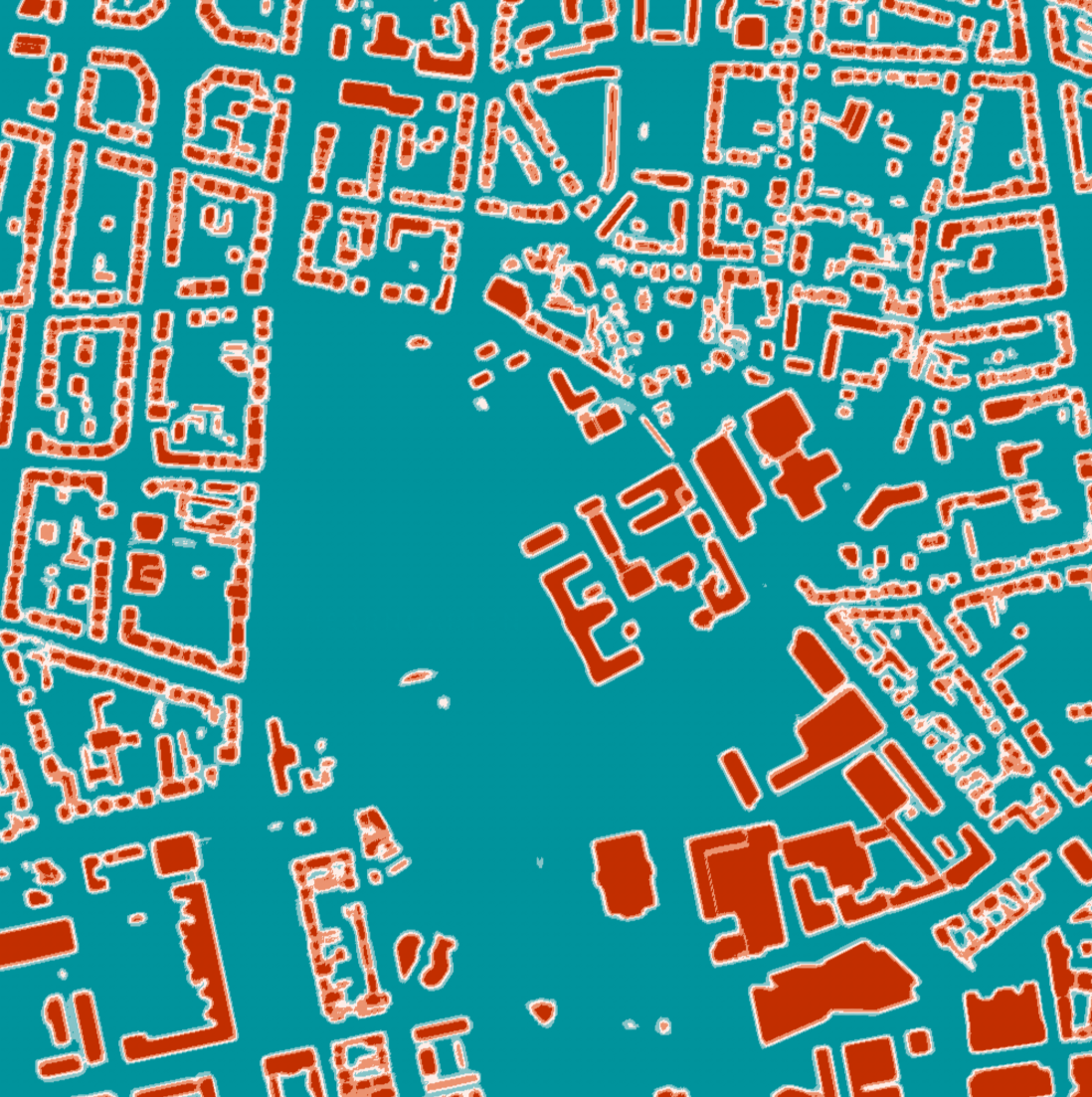}
        \caption{$P(O)$}
    \end{subfigure}
    \caption{
        \textbf{Scalar and vector fields in StaR Maps express uncertainties in spatial relations:}
        Queried road network for which random maps have been generated. (a) and (b) show parameters of a normal distribution that model the distance to the closest road, while (c) models the probability of keeping a distance over 30~m.  
        (d) shows the probability of a location of an agent's navigation space being occupied by buildings. Note a color range from red (low) to blue (high) for (a)-(c) and an inverse color range due to visual clarity for (d).}
    \label{fig:distributional_atoms}
\end{figure*}

%% file: content/4_experiments.tex
\section{Experiments}
\label{sec:results}

\subsection{Setup}

We are employing real-world data that the OpenStreetMap (OSM) community has collectively annotated with semantics such as fine-granular road types, building aspects, and amenities.
However, OSM does not contain uncertainty annotations at this moment.
Hence, for this evaluation, we annotate all map features with the same synthetic uncertainty such that only a translation error $\vec{t}$ is generated from a Gaussian $\mathcal{N}(0, \text{diag}(10m, 10m))$ model.
While this is a stark simplification of the real-world characteristic of such maps, e.g., their heterogeneous accuracy across features, this setup is sufficient to evaluate the creation and application of StaR Maps.  
The implementation employed for these experiments is available as OpenSource software alongside our Probabilistic Mission Design framework \textit{https://github.com/HRI-EU/ProMis}.

\input{figures/distances}
\input{figures/grid_interpolation}
\input{figures/time_error}

\subsection{Creating StaR Maps}

We demonstrate the computation and iterative refinement of StaR Maps.
As a first step, let us take a look at sampling a spatial relation from a single point.
To do so, we generate $N = 50$ maps from the OSM-based UAM as described in Section~\ref{sec:methods} and compute the distance to the respective closest road-type element.
The result for a single location can be seen in Figure~\ref{fig:distances}, where the set of samples and estimated distribution of the spatial relation are portrayed as histogram and Gaussian, respectively.

Now, let us explore how we can efficiently compute from the data contained in a Star Map, e.g., sets of sampled data in $G_{distance, road}$, containing mean and variance information about the \textit{distance} relation as seen in Figure~\ref{fig:distributional_atoms}.
As a first approach, we compute the parameters from a raster of samples and, as a naive choice for the function $v_{distance, road}$, apply linear interpolation to obtain the scalar field of this parameter.
Figure~\ref{fig:grid_interpolation} shows how the chosen grid of samples needs to be appropriately dense to match a high-resolution reference accurately.
Of course, while this strategy can reliably reduce the error, the computational effort grows quadratically, as seen in Figure~\ref{fig:time_error}.
Note that this approach is further limited to the confinements of the convex hull of the samples.

\input{figures/gp_guided}
\input{figures/time_error_gp}

Instead, consider a more advanced choice for $v_{distance, road}$ by fitting a Gaussian Process (GP) on the samples to (i) predict across a continuous space and (ii) guide an incremental refinement based on the GP's confidence.
To do so, we initialize the GP with a random sample of $256$ points.
Then, by predicting the same high-resolution reference of $512 \times 512$ points, we obtain the mean and standard deviation.
By choosing points where the standard deviation is high, meaning low confidence of the GP based on his training data, we sample the spatial relation at those locations and add them to the training set.
This way, the GP yields a higher accuracy and more sample-efficient result than linear interpolation.
Further, considering the GP's confidence across the relevant space, one can decide when to stop the iterative refinements.
Figure~\ref{fig:gp_guided} shows an initial state as well as the prediction, confidence, and remaining error at different stages of refinement.
Further, Figure~\ref{fig:time_error_gp} shows the comparatively low demand in time to gather samples and high prediction quality.

\subsection{Reasoning on StaR Maps}

Beyond representing the statistics of relations between points in space and the semantics within uncertain maps, StaR Maps facilitate a basis for spatial reasoning.
Analogously to prior work~\cite{Kohaut2023}, one can query spatial requests to StaR Maps as a statistical relational knowledge base by employing a hybrid probabilistic, first-order logic.

StaR Maps contain two types of knowledge, represented by categorical and continuous distributions, respectively.
As basic running examples, making up a considerable proportion of public legislative language, we have shown \textit{over} and \textit{distance} as binary predicates between points in space and different environment semantics.
To this end, we show how this vocabulary can be employed for spatial reasoning, i.e., deriving further knowledge beyond what is already encapsulated in the Star Map.

A hybrid probabilistic logic programming language such as presented by Nitti et al.~\cite{nitti} can be employed to represent both categorical and continuous distributions in a declarative modeling language.
More critically, they allow for querying the probability of logical conjunctions and disjunctions within the model.
To demonstrate the application of StaR Maps, we show such a program in Listing~\ref{listing:uam_model}.
For each chosen point $\vec{x}$ within the navigation space, the StaR Map data has been written out in the form of distributional clauses, e.g., for \textit{distance} and \textit{over} relations.
One can further introduce domain knowledge to then query for application dependant probabilities.
While in this example, a simple airspace is defined, Figure~\ref{fig:scenarios} shows a selection of more complex models being evaluated using StaR Map data of the respective locations.

\input{listings/uam_model}

%% file: figures/distances.tex
\begin{figure}
    \centering
    \includegraphics[width=\linewidth]{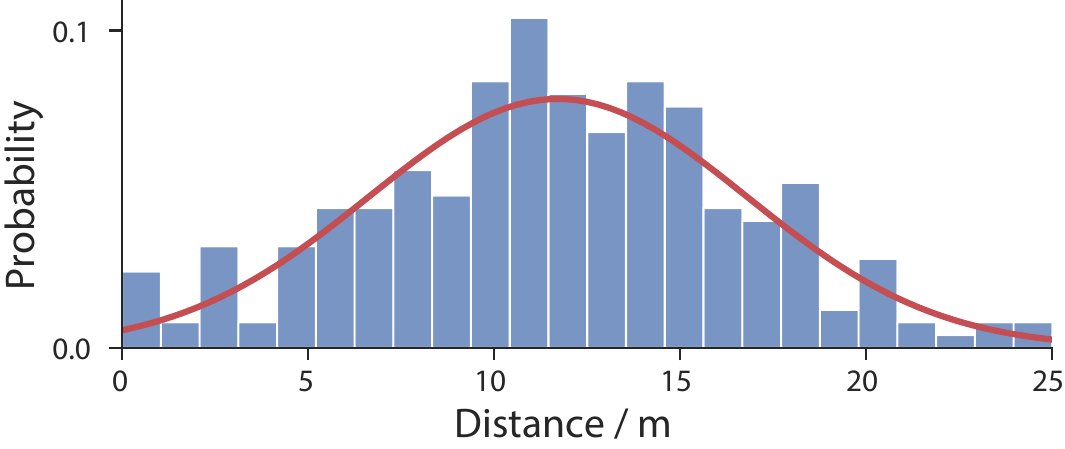}
    \caption{
        \textbf{Parameter estimation of the distance relation:}
        Here, a histogram of the sampling process of the distance to the closest road is shown for a single point.
        From the set of samples we compute mean and standard deviation in order to parameterize the Gaussian that will model the distribution.
    }
    \label{fig:distances}
\end{figure}

%% file: figures/grid_interpolation.tex
\begin{figure}
    \centering
    \includegraphics[width=0.95\linewidth]{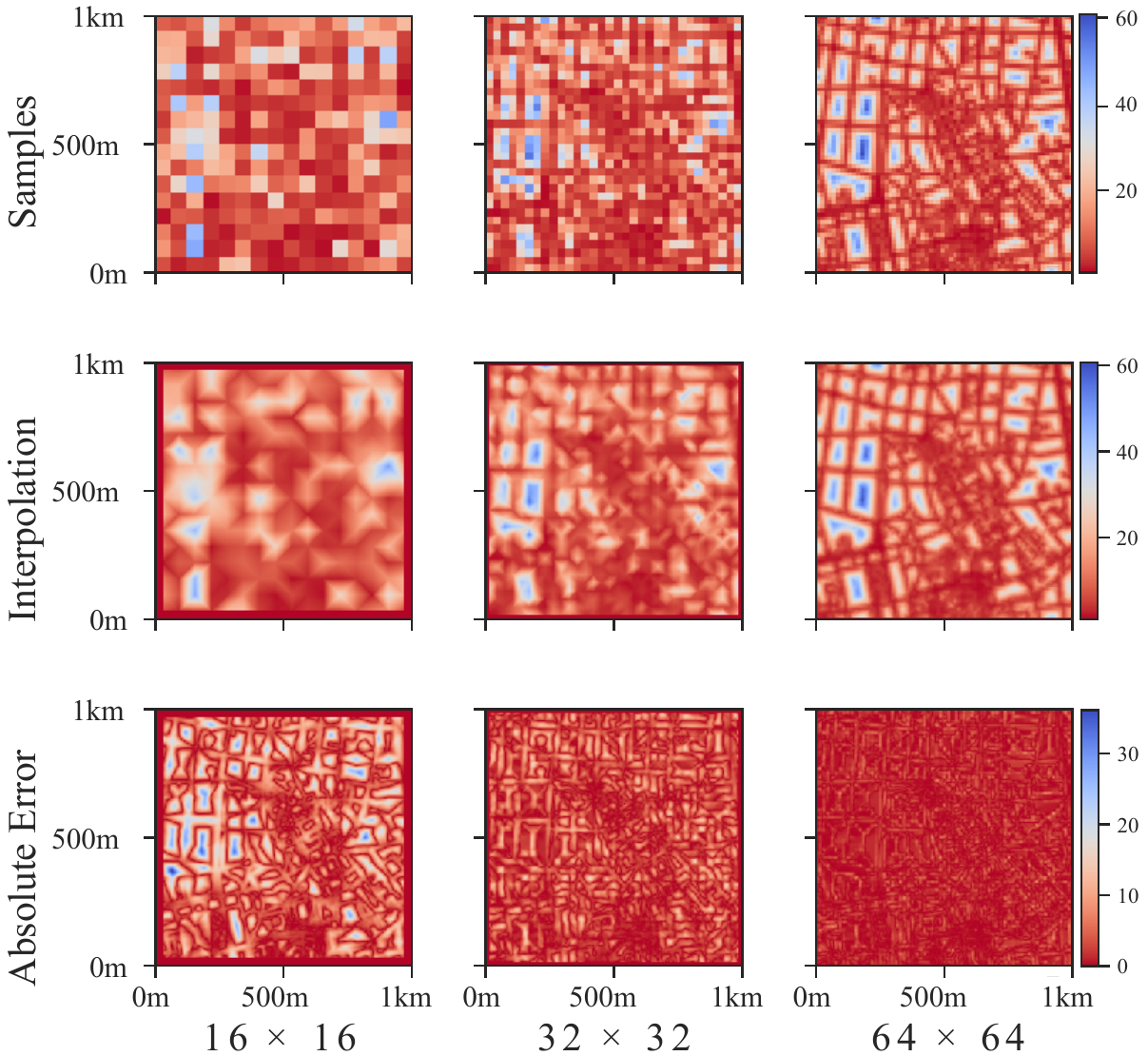}
    \caption{
        \textbf{Interpolation of \textit{distance} mean on raster of samples:}
        Distributing sampling points on a regular grid has the advantage of a uniform interpolation quality across the mapped space.
        A $512 \times 512$ raster of samples was taken as the reference image and compared to interpolated low-resolution data.
    }
    \label{fig:grid_interpolation}
\end{figure}

%% file: figures/time_error.tex
\begin{figure}
    \centering
    \includegraphics[width=\linewidth]{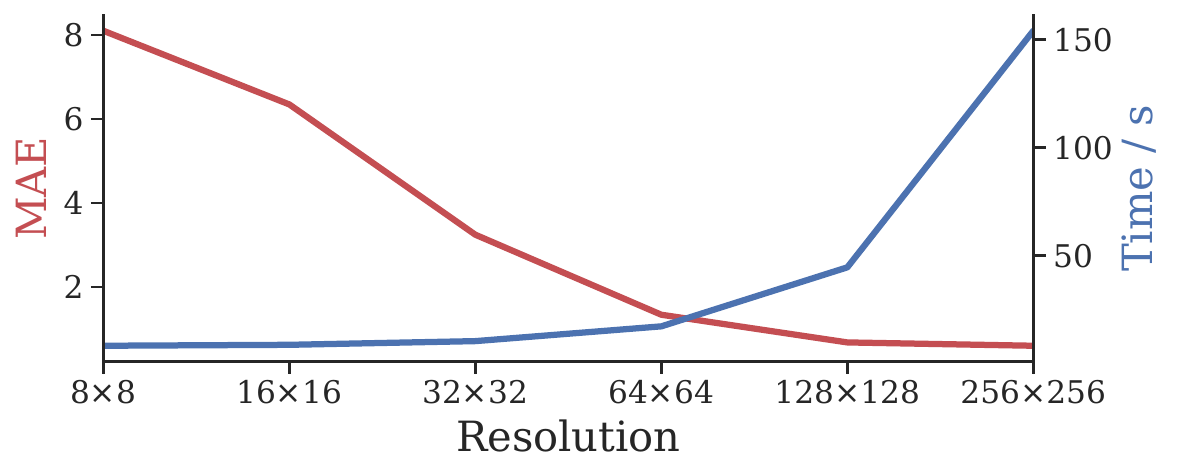}
    \caption{
        \textbf{Interpolation-based StaR Map creation:}
        Analogously to Figure~\ref{fig:grid_interpolation}, we compare the mean absolute error (MAE) of various base resolutions when interpolated and compared to a $512 \times 512$ representation.
        While the error is reduced as expected, computation times grow quadratically due to the sample-grids growing resolution.
43    }
    \label{fig:time_error}
\end{figure}

%% file: figures/gp_guided.tex
\begin{figure}
    \centering
    \includegraphics[width=\linewidth]{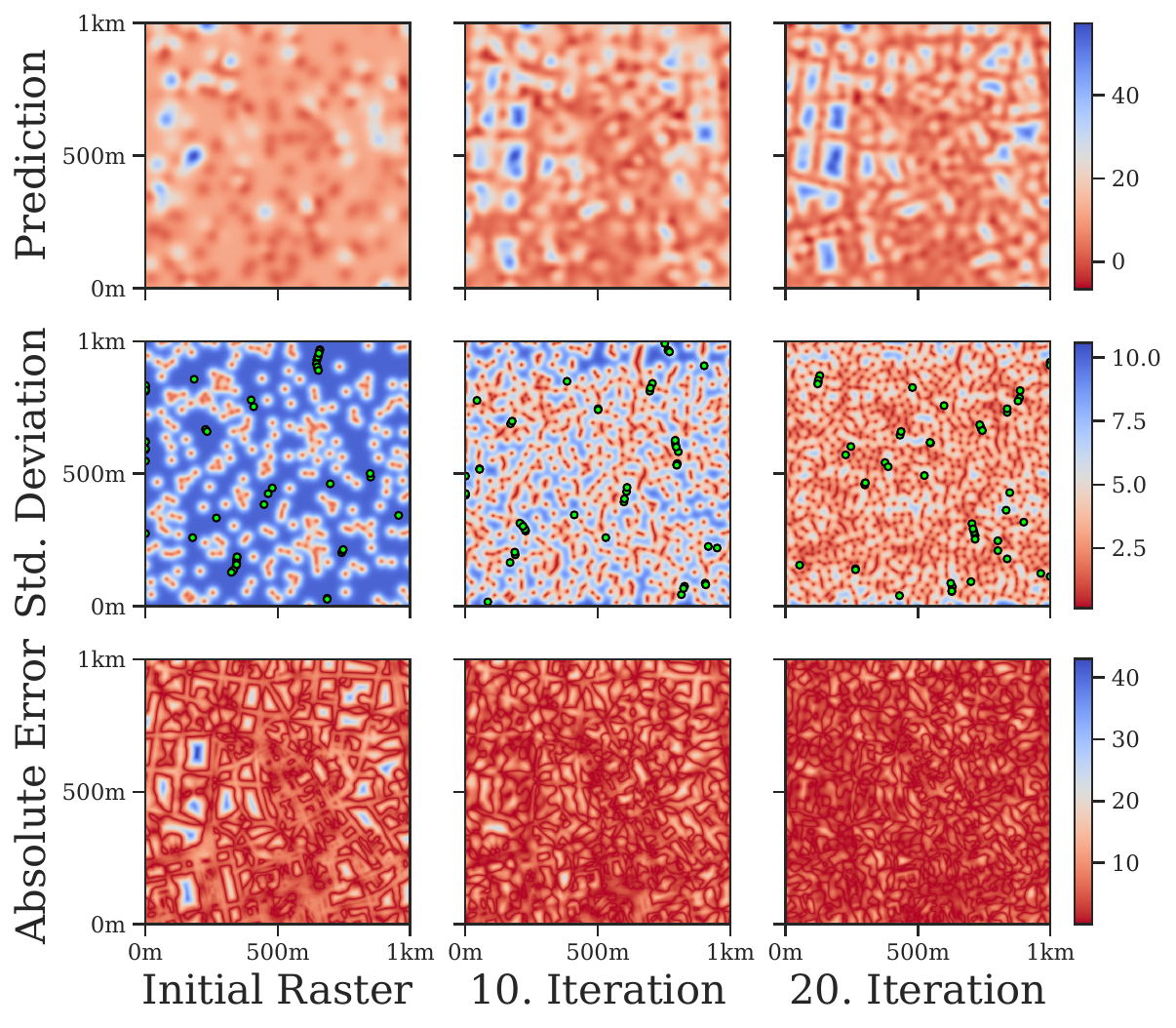}
    \caption{
        \textbf{Gaussian Process (GP) prediction of \textit{distance} mean using confidence guided refinement:}
        The GP's prediction capabilities allow us to obtain mean and standard deviation across the mapped area.
        Sampling at points of low confidence (green dots), we can incrementally refine the GP's training set to capture the spatial relation more accurately.
    }
    \label{fig:gp_guided}
\end{figure}

%% file: figures/time_error_gp.tex
\begin{figure}
    \centering
    \includegraphics[width=\linewidth]{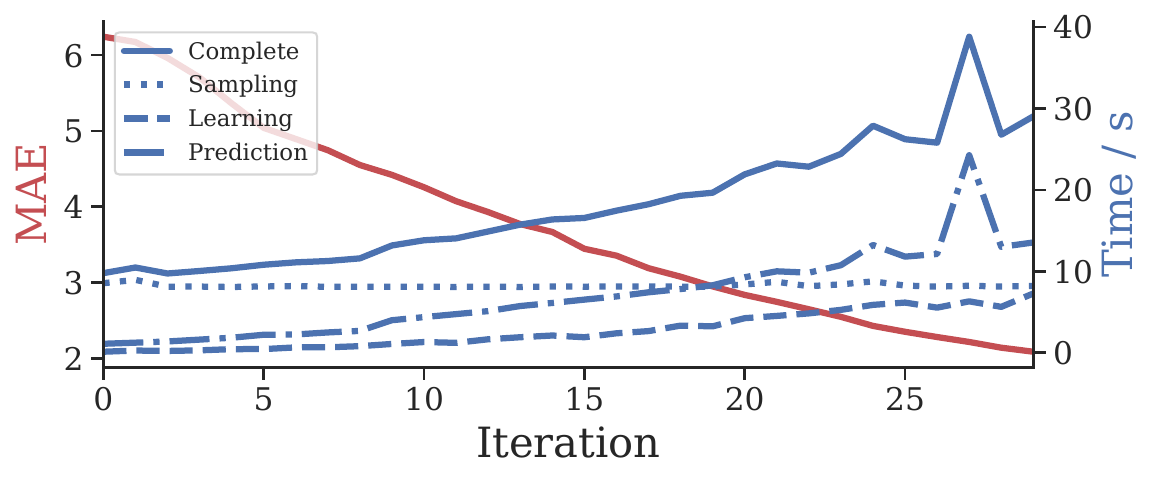}
    \caption{
        \textbf{Gaussian Process (GP) based StaR Map creation:}
        In contrast to grid-based sampling as seen in Figure~\ref{fig:time_error}, the GP incrementally lowers the error while only taking on few more training points at the actually relevant locations.
        Although this approach has advantages, one has to note the necessity of tuning hyper-parameters for a reliable application.
    }
    \label{fig:time_error_gp}
\end{figure}

%% file: listings/uam_model.tex
\begin{listing}
    \centering
    \begin{minted}
    [
        autogobble,
        fontsize=\footnotesize,
    ]{prolog}
    % Quantitative uncertainties, e.g., distance
    distance(x, building) ~ normal(20, 0.5).
    
    % Categorical uncertainties, e.g., over 
    0.9::over(x, primary).
    
    % The queried space and its constraint
    airspace(X) :- over(X, park).
    airspace(X) :- distance(X, road) < 15, 
        distance(X, pilot) < 250.
    \end{minted}
    \caption{
        \textbf{StaR Maps facilitate spatial reasoning:}
        By translating the statistical, relational knowledge of the StaR Map into a hybrid probabilistic logic program, one can reason about spaces where all constraints are likely fulfilled.
        This example defines a simple airspace for an Unmanned Aerial Vehicle by restricting the valid airspace to be over park areas or close to primary roads and the remote pilot. 
    }
    \label{listing:uam_model}
\end{listing}

%% file: content/5_conclusion.tex
\section{Conclusion}
\label{sec:conclusion}

We have presented StaR maps, a novel, hybrid probabilistic and relational environment representation.
StaR Maps facilitate a probabilistic, relational worldview and enable complex reasoning about domain-specific queries in first-order logic.
Further, StaR Maps improves the representation of spatial relations in contrast to prior work~\cite{Kohaut2023} by incorporating a probabilistic regression model on top of the sampling process, leading to confidence-guided refinements and additional information on the data's quality.
We further demonstrated how reasoning on StaR Maps through spatial relations provides a foundation for building rich languages for tasks such as navigation or plan validation.

\input{figures/scenarios}

Future work includes extensions of our spatial relations into many domains, embracing the diversity of multiple modes of mobility and their demands.
While we have shown how to incorporate uncertainties connected to individual map elements and how to obtain the statistics of spatial relations efficiently, guiding the sampling process to intelligently select optimal places for computation and reducing redundancies in the process is essential.

A crucial component of our framework is modeling errors and spatial relations. 
One approach towards online identification of such parameters in the case of road segments has been demonstrated in~\cite{flade2021error}.
However, further investigation into a more general and robust approach is important to gather uncertainty information on a wider range of geographic features.
This task is vital to future map-making and maintenance to lift geographic representations into a probabilistic domain.

Further, it is vital to consider the underlying assumptions of a spatial relation's distribution.
While we have utilized, e.g., Gaussian distributions in our presentation, they are not general tools to capture all data accurately.
In future work, it will be essential to consider what models to employ and to build up a broader collection of spatial relations alongside them.
A proper assessment of these choices' quality will be critical to guide safe and reliable decision-making.

%% file: figures/scenarios.tex
\begin{figure}
    \centering
    \begin{subfigure}{0.39\linewidth}
        \includegraphics[width=\textwidth]{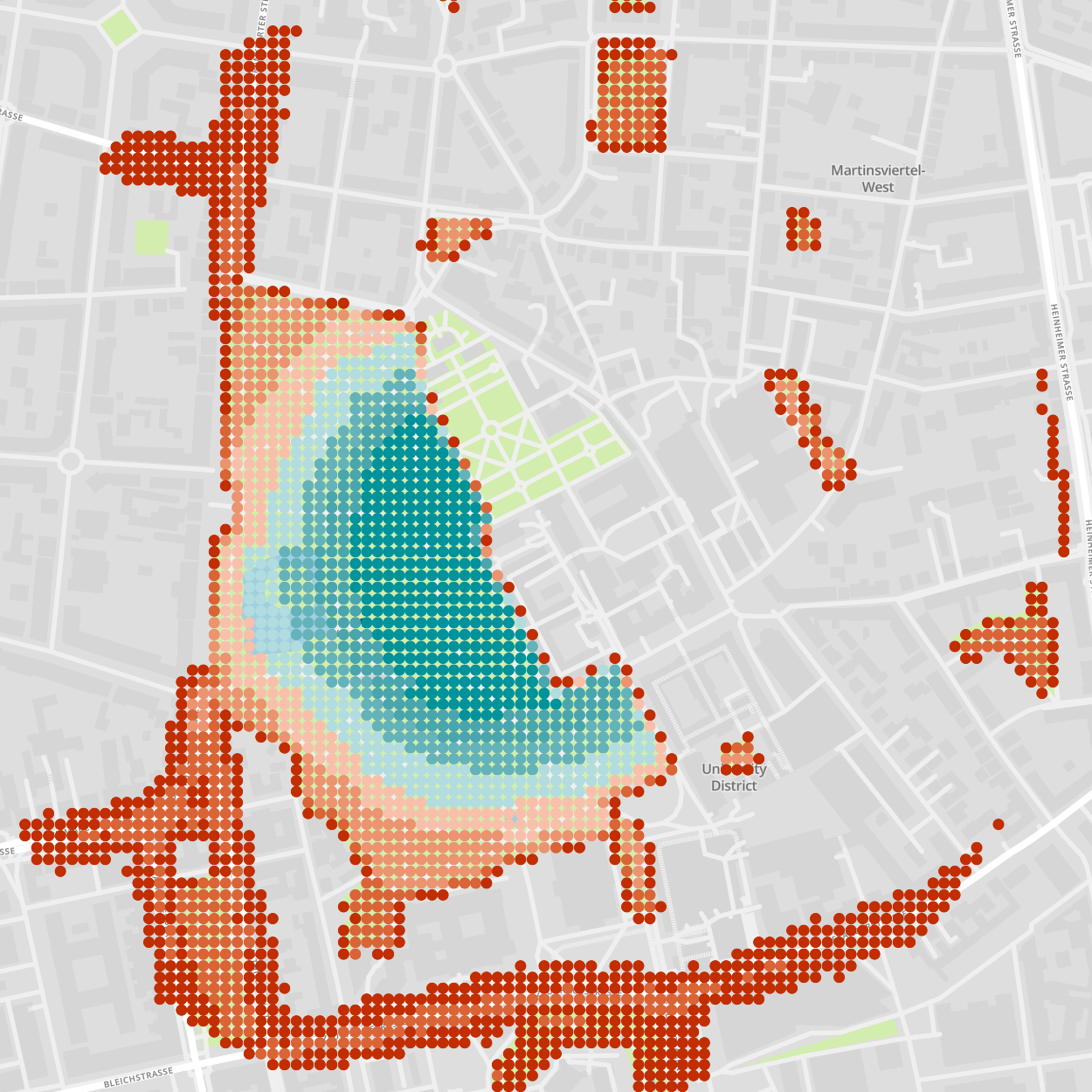}
        \caption{}
    \end{subfigure}
    \begin{subfigure}{0.39\linewidth}
        \includegraphics[width=\textwidth]{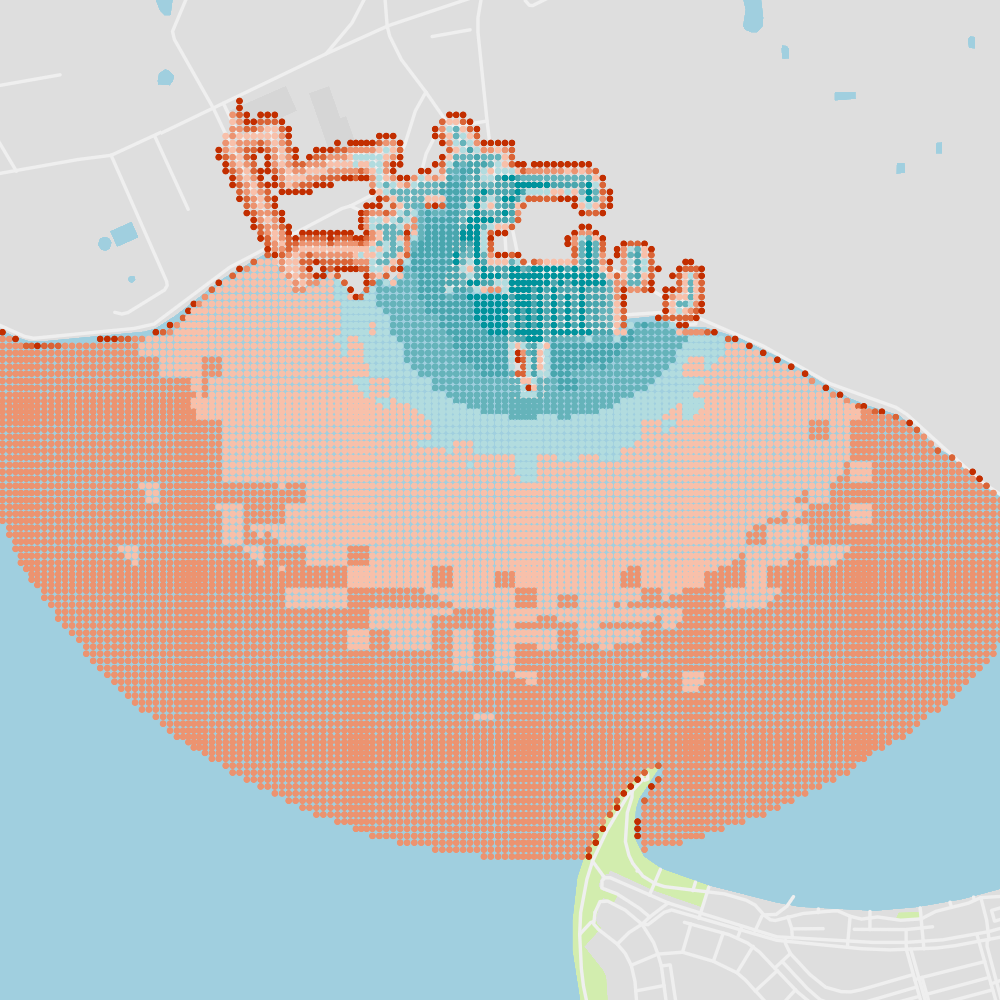}
        \caption{}
    \end{subfigure} \\
    \vspace{0.35em}
    \begin{subfigure}{0.39\linewidth}
        \includegraphics[width=\textwidth]{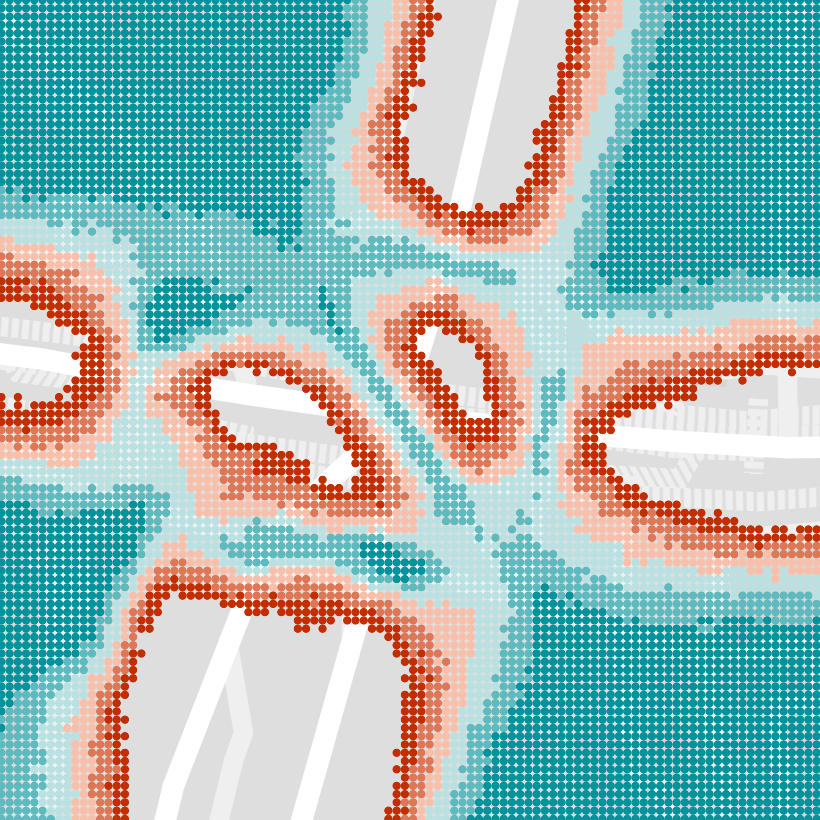}
        \caption{}
    \end{subfigure}
    \begin{subfigure}{0.39\linewidth}
        \includegraphics[width=\textwidth]{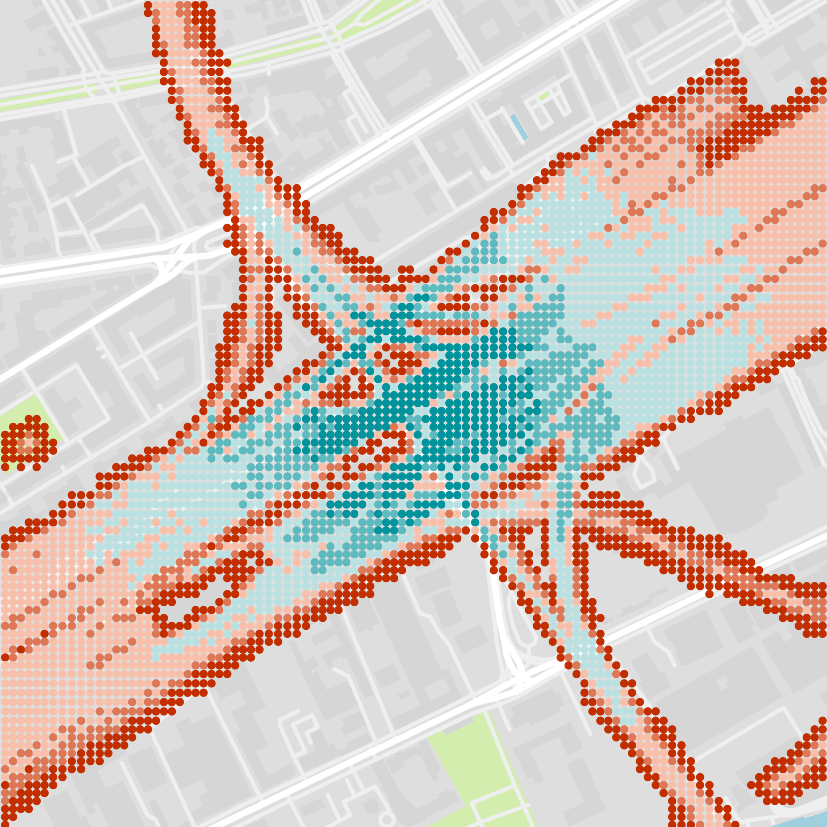}
        \caption{}
    \end{subfigure}
    \caption{
        \textbf{Reasoning on StaR Maps:}
        Through StaR Maps' vocabulary of probabilistic predicates over the navigation space, they facilitate complex semantic queries.
        Across a variety of environments, namely (a) an urban city park, (b) a bay area, (c) a major junction with a pedestrian crossing, and (d) a central train station for cargo and passenger transport.
    }
    \label{fig:scenarios}
\end{figure}

%% file: content/6_acknowledgment.tex
\section*{Acknowledgments}

The Eindhoven University of Technology authors received support from their Department of Mathematics and Computer Science and the Eindhoven Artificial Intelligence Systems Institute.
Map data \copyright~OpenStreetMap contributors, licensed under the Open Database License (ODbL) and available from https://www.openstreetmap.org. Map styles \copyright~Mapbox, licensed under the Creative Commons Attribution 3.0 License (CC BY 3.0) and available from https://github.com/mapbox/mapbox-gl-styles.